\documentclass[10pt]{article}

\pdfoutput=1 %

\usepackage[utf8]{inputenc}
\usepackage{amsmath}
\usepackage{amssymb}
\usepackage{amsfonts}
\usepackage{pifont}

\usepackage[backend=bibtex,
style=numeric,
bibencoding=ascii,
maxbibnames=99,
]{biblatex}
\usepackage[compact]{titlesec}
\usepackage[utf8]{inputenc} %
\usepackage[T1]{fontenc}    %

\usepackage[breaklinks=true,colorlinks,citecolor=black,bookmarks=false]{hyperref}
\hypersetup{
	colorlinks=true,
	linkcolor=blue,
	filecolor=magenta,      
	urlcolor=blue,
	citecolor=black,
  pdfinfo={
    Title={Unsolved Problems in ML Safety},
    Author={Dan Hendrycks and Nicholas Carlini and John Schulman and Jacob Steinhardt},
    Subject={ml safety, ai safety, trustworthy ml},
    Keywords={robustness, proxy gaming, transparency, machine ethics, value learning, wellbeing, uncertainty, deception, existential risk, anomaly detection, forecasting, calibration, cybersecurity}
  }
}

\usepackage{url}            %
\usepackage{booktabs}       %
\usepackage{amsfonts}       %
\usepackage{nicefrac}       %
\usepackage{microtype}      %

\usepackage{times}
\usepackage{epsfig}
\usepackage{wrapfig}

\usepackage[utf8]{inputenc} %
\usepackage{amsmath, amssymb}
\usepackage[T1]{fontenc}    %
\usepackage{url}            %
\usepackage{amsfonts}       %
\usepackage{amsbsy}
\usepackage{microtype}      %
\usepackage{wrapfig}
\usepackage{pdfpages}
\usepackage{caption}
\usepackage{subcaption}
\usepackage{hhline}
\usepackage{multirow}
\usepackage{tabularx}
\newcolumntype{Y}{>{\centering\arraybackslash}X}
\usepackage{arydshln}
\usepackage{multirow}
\usepackage{makecell}

\usepackage{mathtools}
\usepackage[utf8]{inputenc}
\usepackage{stfloats}
\usepackage{float}
\usepackage{enumitem}
\usepackage{titling} %

\definecolor{rightgreen}{RGB}{0,154,24}

\usepackage{cleveref}
\usepackage{appendix}
\crefname{appsec}{Appendix}{Appendices}

\usepackage[most]{tcolorbox}

\usepackage{spverbatim}

\makeatletter
\newcommand*\bigcdot{\mathpalette\bigcdot@{.5}}
\newcommand*\bigcdot@[2]{\mathbin{\vcenter{\hbox{\scalebox{#2}{$\m@th#1\bullet$}}}}}
\makeatother

\makeatletter
\newcommand{\printfnsymbol}[1]{%
  \textsuperscript{\@fnsymbol{#1}}%
}
\makeatother

\newcommand{\reviewer}[3]{
	\expandafter\newcommand\csname #1\endcsname[1]{
		\textcolor{#3}{[#2: ##1]}
	}
}
\reviewer{someone}{Someone}{red}
\definecolor{neonpurple}{rgb}{0.3,0,1}
\reviewer{dan}{Dan}{neonpurple}
\reviewer{js}{Jacob}{blue}
\reviewer{nicholas}{Nicholas}{red}
\reviewer{tom}{Tom}{green}

\title{
\vspace{-15pt}
\rule[0.4cm]{\textwidth}{2pt}
{\bf Unsolved Problems in ML Safety}
\rule{\textwidth}{2pt} 
}
\date{}

\author{\textbf{Dan Hendrycks}\\
UC Berkeley\\
\and
\textbf{Nicholas Carlini}\\
Google\\
\and
\textbf{John Schulman}\\
OpenAI\\
\and
\textbf{Jacob Steinhardt}\\
UC Berkeley\\
}

\begin{document}

\maketitle

\vspace*{-25pt}
\begin{abstract}%
\normalsize
Machine learning (ML) systems are rapidly increasing in size, are acquiring new capabilities, and are increasingly deployed in high-stakes settings. As with other powerful technologies, safety for ML should be a leading research priority. In response to emerging safety challenges in ML, such as those introduced by recent large-scale models, we provide a new roadmap for ML Safety and refine the technical problems that the field needs to address.
We present four problems ready for research, namely withstanding hazards (``Robustness''), identifying hazards (``Monitoring''), steering ML systems (``Alignment''), and reducing deployment hazards (``Systemic Safety''). Throughout, we clarify each problem's motivation and provide concrete research directions.\looseness=-1
\end{abstract}

\vspace{7pt}
\section{Introduction}
\vspace{3pt}

As machine learning (ML) systems are deployed in high-stakes environments, such as medical settings \cite{Rajpurkar2017CheXNetRP}, roads \cite{teslaaiday}, and command and control centers \cite{gide3}, unsafe ML systems may result in needless loss of life. %
Although researchers recognize that safety is important \cite{asilomar,Amodei2016ConcretePI}, 
it is often unclear what problems to prioritize or how to make progress.
We identify four problem areas that would help make progress on ML Safety: robustness, monitoring, alignment, and systemic safety. While some of these, such as robustness, are long-standing challenges, the success and emergent capabilities of modern ML systems necessitate new angles of attack.\looseness=-1

We define ML Safety research as ML research aimed at making the adoption of ML more beneficial, with emphasis on long-term and long-tail  risks.
We focus on cases where greater capabilities can be expected to decrease safety, or where ML Safety problems are otherwise poised to become more challenging in this decade.
For each of the four problems, after clarifying the motivation, we discuss possible research directions that can be started or continued in the next few years. 
First, however, we motivate the need for ML Safety research.\looseness=-1

We should not procrastinate on safety engineering. In a report for the Department of Defense, Frola and Miller \cite{Frola1984SystemSI} observe that approximately $75\%$ of the most critical decisions that determine a system's safety occur early in development \cite{Leveson2012EngineeringAS}. If attention to safety is delayed, its impact is limited, as unsafe design choices become deeply embedded into the system.
The Internet was initially designed as an academic tool with neither safety nor security in mind \cite{denardis2007history}.
Decades of security patches later, security measures are still incomplete and increasingly complex. 
A similar reason for starting safety work now is that relying on experts to test safety solutions is not enough---solutions must also be age tested.
The test of time is needed even in the most rigorous of disciplines. A century before the four color theorem was proved, Kempe's peer-reviewed proof went unchallenged for years until, finally, a flaw was uncovered \cite{Heawood1949MapColourT}. Beginning the research process early allows for more prudent design and more rigorous testing.
Since nothing can be done both hastily and prudently \cite{syrus1856moral}, postponing %
machine learning safety research increases the likelihood of accidents.

Just as we cannot procrastinate, we cannot rely exclusively on previous hardware and software %
engineering practices to create safe ML systems.
In contrast to typical software, ML control flows are specified by inscrutable weights learned by gradient optimizers rather than programmed with explicit instructions and general rules from humans.
They are trained and tested pointwise using specific cases, which has limited effectiveness at improving and assessing an ML system's completeness and coverage.
They are fragile, rarely correctly handle all test cases, and cannot become error-free with short code patches \cite{sculley2015hidden}. They exhibit neither modularity nor encapsulation, making them far less intellectually manageable and making causes of errors difficult to localize. They frequently demonstrate properties of self-organizing systems such as spontaneously emergent capabilities \cite{Brown2020LanguageMA,caron2021emerging}. They may also be more agent-like and tasked with performing open-ended actions in arbitrary complex environments. %
Just as, historically, safety methodologies developed for electromechanical hardware \cite{StamatisFailureMA} did not generalize to the new issues raised by software, we should expect 
software safety methodologies not to generalize to the new complexities and hazards of ML.

We also cannot solely rely on economic incentives and regulation to shepherd competitors into developing safe models.
The competitive dynamics surrounding ML's development may pressure companies and regulators to take shortcuts on safety. 
Competing corporations often prioritize minimizing development costs and being the first to the market over providing the safest product.
For example, Boeing developed the 737 MAX with unsafe design choices to keep pace with its competitors; and as a direct result of taking shortcuts on safety and pressuring inspectors, Boeing's defective model led to two crashes across a span of five months that killed 346 people \cite{sumwalt2019assumptions,Folkert2021,Ky2021}.
Robust safety regulation is almost always developed only after a catastrophe---a common saying in aviation is that ``aviation regulations are written in blood.'' While waiting for catastrophes to spur regulators
can reduce the likelihood of repeating the same failure, %
this approach cannot prevent catastrophic events from occurring in the first place. Regulation efforts may also be obstructed by lobbying or by the spectre of lagging behind international competitors who may build superior ML systems. Consequently, companies and regulators may be pressured to deprioritize safety.\looseness=-1

\begin{figure*}[t]
    \centering
    \includegraphics[width=\textwidth]{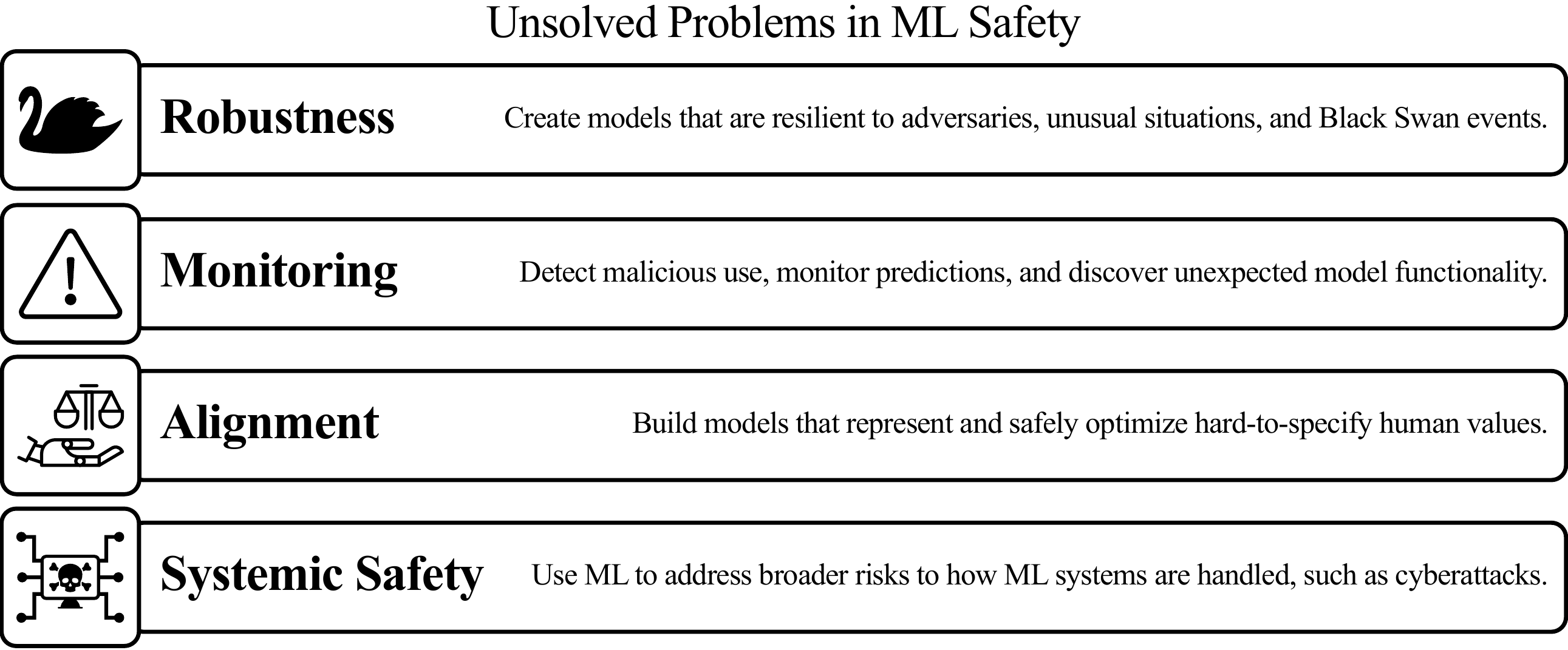}
    \label{fig:splash}
    \vspace{-10pt}
\end{figure*}

These sources of hazards---starting safety research too late, novel ML system complexities, and competitive pressure---may result in deep design flaws. However, a strong safety research community can drive down these risks.
Working on safety proactively builds more safety into systems during the critical early design window.
This could help reduce the cost of building safe systems and reduce the pressure on companies to take shortcuts on safety.
If the safety research community grows, it can help handle the spreading multitude of hazards that continue to emerge as ML systems become more complex.
Regulators can also prescribe higher, more actionable, and less intrusive standards if the community has created ready-made safety solutions.

When especially severe accidents happen, everyone loses.
Severe accidents can cast a shadow that creates unease and precludes humanity from realizing ML's benefits. Safety engineering for powerful technologies is challenging, as the Chernobyl meltdown, the Three Mile Island accident, and the Space Shuttle Challenger disaster have demonstrated. However, done successfully, work on safety can improve the likelihood that essential technologies operate reliably and benefit humanity.%

\section{Robustness}\label{sec:robustness}

\begin{figure}[H]
    \centering
    \includegraphics[width=0.75\textwidth]{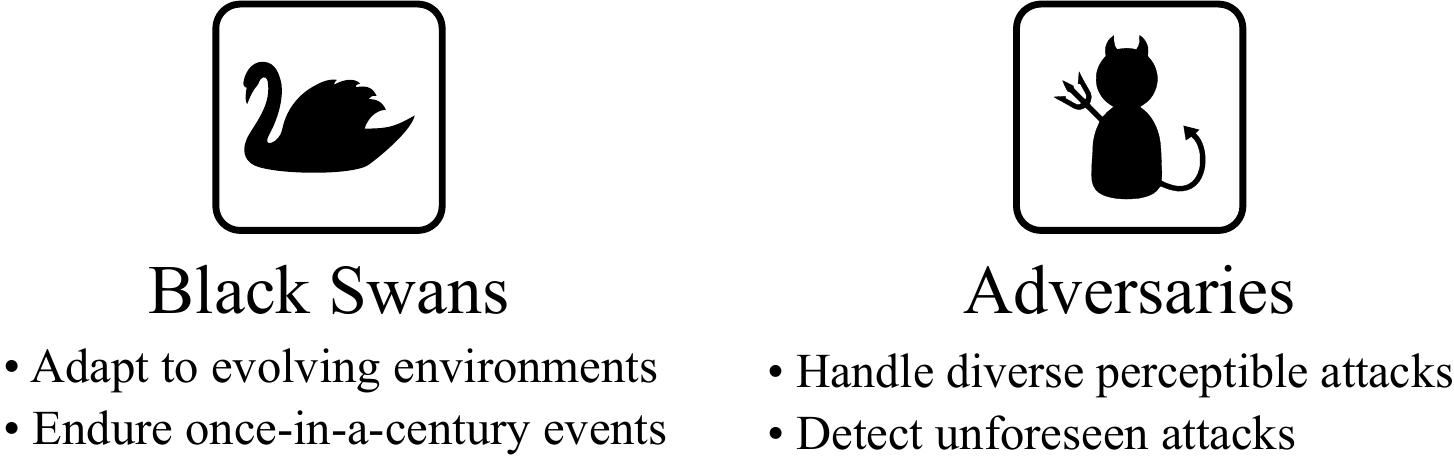}
    \caption{Robustness research aims to build systems that endure extreme, unusual, or adversarial events.%
}
    \label{fig:robustness}
\end{figure}

\subsection{Black Swan and Tail Risk Robustness}\label{sec:longtail}

\paragraph{Motivation.} To operate in open-world high-stakes environments, machine learning systems will need to endure unusual events and tail risks. However, current ML systems are often brittle in the face of real-world complexity and unknown unknowns. In the 2010 Flash Crash \cite{Kirilenko2011TheFC}, automated trading systems unexpectedly overreacted to market aberrations, created a feedback loop, and wiped away a trillion dollars of stock value in a matter of minutes. This demonstrates that computer systems can both create and succumb to long tail events.

Long tails continue to thwart modern ML systems such as autonomous vehicles. This is because some of the most basic concepts in the real world are long tailed, such as stop signs, where a model error can directly cause a crash and loss of life. Stop signs may be titled, occluded, or represented on an LED matrix; sometimes stop signs should be disregarded, for example when held upside down by a traffic officer, on open gates, on a shirt, on the side of bus, on elevated toll booth arms, and so on. Although these long tail events are rare, they are
\begin{wrapfigure}{r}[0.01\textwidth]{.38\textwidth}%
	\vspace{-8pt}%
    ``Things that have never happened before happen all the time.''\hfill\emph{Scott D.\ Sagan}%
	\vspace{-10pt}%
\end{wrapfigure}
extremely impactful \cite{Taleb2020StatisticalCO} and can cause ML systems to crash. Leveraging existing massive datasets is not enough to ensure robustness, as models trained with Internet data and petabytes of task-specific driving data still are not robust to long tail road scenarios \cite{teslaaiday}. This decades-long challenge is only a preview of the more difficult problem of handling tail events in environments that are beyond a road's complexity.\looseness=-1 %

Long-tail robustness is unusually challenging today and may become even more challenging. Long-tail robustness also requires more than human-level robustness; the 2008 financial crisis and COVID-19 have shown that even groups of humans have great difficulty mitigating and overcoming these rare but extraordinarily impactful long tail events. Future ML systems will operate in environments that are broader, larger-scale, and more highly connected with more feedback loops, paving the way to more extreme events \cite{Mitzenmacher2003ABH} than those seen today.\looseness=-1

While there are incentives to make systems partly robust, systems tend not to be incentivized nor designed for long tail events outside prior experience, even though Black Swan events are inevitable 
\cite{usplanning}. To reduce the chance that ML systems will fall apart in settings dominated by rare events, systems must be \emph{unusually} robust.\looseness=-1

\paragraph{Directions.} In addition to existing robustness benchmarks \cite{hendrycks2019robustness,Koh2021WILDSAB,hendrycks2021many}, researchers could create more environments and benchmarks to stress-test systems, find their breaking points, and determine whether they will function appropriately in potential future scenarios.
These benchmarks could include new, unusual, and extreme distribution shifts and long tail events, especially ones that are challenging even for humans. Following precedents from industry \cite{teslaaiday,waymo}, benchmarks could include artificial simulated data that capture structural properties of real long tail events. Additionally, benchmarks should focus %
on ``wild'' distribution shifts that cause large accuracy drops over ``mild'' shifts \cite{Mandelbrot2004TheMO}.\looseness=-1  %

Robustness work could also move beyond classification and consider \emph{competent errors} where agents misgeneralize and execute wrong routines, such as an automated digital assistant knowing how to use a credit card to book flights, but choosing the wrong destination \cite{Koch2021ObjectiveRI,Hubinger2019RisksFL}.
Interactive environments \cite{Cobbe2019QuantifyingGI} could simulate qualitatively
distinct random shocks that irreversibly shape the environment's future evolution. Researchers could also create environments where ML system outputs affect their environment and create feedback loops.%

Using such benchmarks and environments, researchers could improve ML systems to withstand Black Swans \cite{Taleb2007TheBS, Taleb2020StatisticalCO}, long tails, and structurally novel events. %
The performance of many ML systems is currently largely shaped by data and parameter count, so future research could work on creating highly unusual but helpful data sources. 
The more experience a system has with unusual future situations, even ones not well represented in typical training data, the more robust it can be.
New data augmentation techniques \cite{hendrycks2021pixmix,hendrycks2020augmix} and other sources of simulated data could create inputs that are not easy or possible to create naturally. %

Since change is a part of all complex systems, and since not everything can be anticipated during training, models will also need to adapt to an evolving world and improve from novel experiences \cite{Mummadi2021TestTimeAT,Wang2021TentFT,Taleb2012AntifragileTT}. %
Future adaptation methods could improve a system's ability to adapt quickly.
Other work could defend adaptive systems against poisoned data encountered during deployment \cite{tay}.

\subsection{Adversarial Robustness}\label{sec:advex}

\paragraph{Motivation.} We now turn from %
unpredictable accidents to carefully crafted and deceptive threats. 
Adversaries can easily manipulate vulnerabilities in ML systems and cause them to make mistakes \cite{biggio2013evasion,szegedy2013intriguing}. %
For example, systems may use neural networks to detect intruders \cite{Ahmad2021NetworkID} or malware \cite{Suciu2019ExploringAE}, but if adversaries can modify their behavior to deceive and bypass detectors, the systems will fail.
While defending against adversaries might seem to be a straightfoward problem,
defenses are currently struggling to keep pace with attacks \cite{Athalye2018ObfuscatedGG,Tramr2020OnAA}, and much research is needed to discover how to fix these longstanding weaknesses.

\paragraph{Directions.}  We encourage research on adversarial robustness to focus on broader robustness definitions.
Current research largely focuses on the problem of ``$\ell_p$ adversarial robustness,'' \cite{Madry2018TowardsDL, carlini2017towards} where an adversary attempts to induce a misclassification but can only perturb inputs subject to a small $p$-norm constraint.
While research on simplified problems helps drive progress, researchers may wish to avoid focusing too heavily on any one particular simplification.

To study adversarial robustness more broadly \cite{Gilmer2018MotivatingTR}, researchers could consider attacks that are perceptible \cite{Poursaeed2021RobustnessAG} or whose specifications are not known beforehand \cite{Kang2019TestingRA,Laidlaw2021PerceptualAR}.
For instance, there is no reason that an adversarial malware sample would have to be imperceptibly similar to some other piece of benign software---as long as the detector is evaded, the attack has succeeded \cite{pierazzi2020intriguing}. Likewise, copyright detection systems cannot reasonably assume that attackers will only construct small $\ell_p$ perturbations to bypass the system, as attackers may rotate the adversarially modified image \cite{engstrom2018rotation} or apply otherwise novel distortions \cite{Gilmer2018MotivatingTR} to the image. 

While many effective attacks assume full access to a neural network, sometimes assuming limited access is more realistic.
Here, adversaries can feed in examples to an ML system and receive the system's outputs, but they do not have access to the intermediate ML system computation \cite{brendel2017decision}.
If a blackbox ML system is not publicly released and can only be queried, it may be possible to practically defend the system against zero-query attacks \cite{Tramr2018EnsembleAT} or limited-query attacks \cite{Chen2019StatefulDO}.

On the defense side, further underexplored assumptions are that systems have multiple sensors or that systems can adapt.
Real world systems, such as autonomous vehicles, have multiple cameras. Researchers could exploit information from these different sensors and find inconsistencies in adversarial images in order to constrain and box in adversaries \cite{Xiao2018CharacterizingAE}. Additionally, while existing ML defenses are typically static, future defenses could evolve during test time to combat adaptive adversaries \cite{Wang2021FightingGW}.\looseness=-1

Future research could do more work toward creating models with adversarially robust representations \cite{Croce2020RobustBenchAS}.
Researchers could enhance data for adversarial robustness by simulating more data \cite{Zhu2021TowardsUT}, augmenting data \cite{Rebuffi2021FixingDA}, repurposing existing real data \cite{Carmon2019UnlabeledDI,Hendrycks2019UsingPC}, and extracting more information from available data \cite{Hendrycks2019UsingSL}. Others could create architectures that are more adversarially robust \cite{Xie2020SmoothAT}. Others could improve adversarial training methods \cite{Wu2020AdversarialWP} and find better losses \cite{Zhang2019TheoreticallyPT,Tack2021ConsistencyRF}.
Researchers could improve adversarial robustness certifications \cite{Raghunathan2018CertifiedDA,lecuyer2019certified,Cohen2019CertifiedAR}, so that models have verifiable adversarial robustness.

It may also be possible to unify the areas of adversarial robustness and robustness to long-tail and unusual events. %
By building systems to be robust to adversarial worst-case environments, they may also be made more robust to random-worse-case environments \cite{Anderson1995ProgrammingSC,NAE}. 
To study adversarial robustness on unusual inputs, researchers could also try detecting adversarial anomalies \cite{Bitterwolf2020CertifiablyAR,NAE} or assigning them low confidence \cite{Stutz2020ConfidenceCalibratedAT}.

\section{Monitoring}

\begin{figure}[H]
    \centering
    \vspace{-10pt}
    \includegraphics[width=0.9\textwidth]{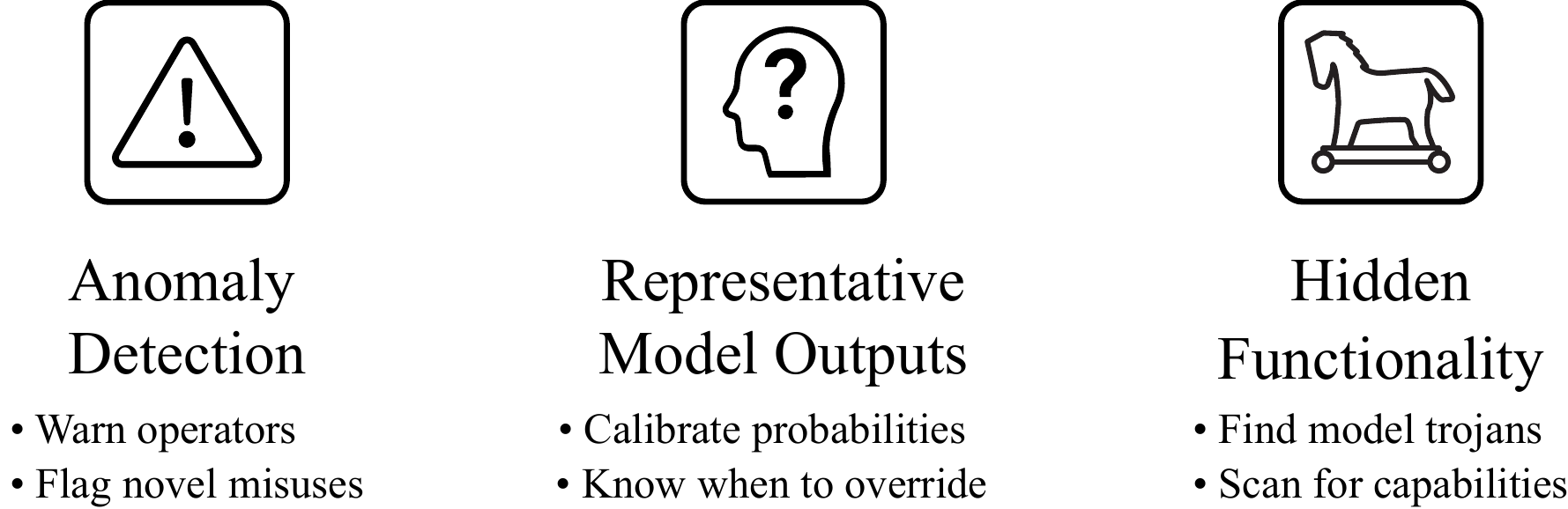}
    \caption{Monitoring research aims to identify hazards, inspect models, and help human ML system operators.}
    \vspace{-2pt}
    \label{fig:monitoring}
\end{figure}

\subsection{Identifying Hazards and Malicious Use With Anomaly Detection}\label{sec:anom}

\paragraph{Motivation.} Deploying and monitoring powerful machine learning systems will require high caution, similar to the caution observed for modern nuclear power plants, military aircraft carriers, air traffic control, and other high-risk systems. These complex and hazardous systems are now operated by high reliability organizations (HROs) which are relatively successful at avoiding catastrophes \cite{Dietterich2018RobustAI}. For safe deployment, future ML systems may be operated by HROs. %
Anomaly detectors are a crucial tool for these organizations since they can warn human operators of potential hazards \cite{hroanomaly}.
For detectors to be useful, research must strive to create detectors with high recall and a low false alarm rate in order to prevent alarm fatigue \cite{cvach2012monitor}.\looseness=-1

Separately, anomaly detection is essential in detecting malicious uses of ML systems \cite{Brundage2018TheMU}. Malicious users are incentivized to use novel strategies, as familiar misuse strategies are far easier to identify and prevent compared to unfamiliar ones. Malicious actors may eventually repurpose ML systems for social manipulation \cite{Buchanan2021Lies}, for assisting research on novel weapons \cite{Bostrom2019TheVW}, or for cyberattacks \cite{Buchanan2020Cyber}. When such anomalies are detected, the detector can trigger a fail-safe policy in the system and also flag the example for human intervention. However, detecting malicious anomalous behavior could become especially challenging when malicious actors utilize ML capabilities to try to evade detection.
Anomaly detection is integral not just for promoting reliability but also for preventing novel misuses.

\paragraph{Directions.} Anomaly detection is actively studied in research areas such as out-of-distribution detection \cite{Hendrycks2017ABF}, open-set detection \cite{Bendale2016TowardsOS}, and one-class learning \cite{Tack2020CSIND, Hendrycks2019UsingSL}, but many challenges remain. The central challenge is that existing methods for representation learning have difficulty discovering representations that work well for previously unseen anomalies.  %
One of the symptoms of this problem is that anomaly detectors for large-scale images still cannot reliably detect that previously unseen random noise is anomalous \cite{Hendrycks2019DeepAD}. %
Moreover, there are many newer settings that require more study, such as detecting distribution shifts or changes to the environment \cite{Danesh2021OutofDistributionDD}, as well developing detectors that work in real-world settings such as intrusion detection, malware detection, and biosafety.

Beyond just detecting anomalies, high reliability organizations require candidate explanations of how an anomaly came to exist \cite{hroanomaly,Siddiqui2019}. To address this, detectors could help identify the origin or location of an anomaly \cite{Besnier2021TriggeringFO}. Other work could try to help triage anomalies and determine whether an anomaly is just a negligible nuisance or is potentially hazardous.

\subsection{Representative Model Outputs}\label{sec:honest}

\subsubsection{Calibration}
\paragraph{Motivation.}
Human monitors need to know when to trust a deployed ML system or when to override it. %
If they cannot discern when to trust and when to override, humans may unduly defer to models and cede too much control. If they can discern this, they can prevent many model hazards and failure modes.

To make models more trustworthy, they should accurately assess their domain of competence \cite{Gil2019A2C}---the set of inputs they are able to handle.
Models can convey the limits of their competency by expressing their uncertainty.
However, model uncertainties are not representative, and they are often overconfident \cite{Guo2017}. To address this, models could become more calibrated.  If a model is perfectly calibrated and predicts a ``$70\%$ chance of rain,'' then when it makes that prediction, $70\%$ of the time it will rain. Calibration research makes model prediction probabilities more representative of a model's overall behavior, provides monitors with a clearer impression of their understanding, and helps monitors weigh model decisions.

\paragraph{Directions.} To help models express their domain of competence in a more representative and meaningful way, researchers could further improve model calibration on typical testing data \cite{Guo2017,Nguyen2015PosteriorCA,Lakshminarayanan2017SimpleAS,Kumar2019VerifiedUC,Zaidi2020NeuralES,Kuleshov2018,Kull2019,Luo2021}, though the greater challenge is calibration on testing data that is unlike the training data \cite{Ovadia2019CanYT}. 
Future systems could communicate their uncertainty with language. For example, they could express decomposed probabilities with contingencies such as ``event $A$ will occur with $60\%$ probability assuming event $B$ also occurs, and with $25\%$ probability if event $B$ does not.'' %
To extend calibration beyond single-label outputs, researchers could take models that generate diverse sentence and paragraph answers and teach these models to assign calibrated confidences to their generated free-form answers.\looseness=-1

\subsubsection{Making Model Outputs Honest and Truthful}

\paragraph{Motivation.} Human monitors can more effectively monitor models if they produce outputs that accurately, honestly, and faithfully \cite{Gilpin2018ExplainingEA} represent their understanding or lack thereof.
However, current language models do not accurately represent their understanding and do not provide faithful explanations. They generate empty explanations that are often surprisingly fluent and grammatically correct but nonetheless entirely fabricated. These models generate distinct explanations when asked to explain again, generate more misconceptions as they become larger \cite{owain2021}, and sometimes generate worse answers when they know how to generate better answers \cite{Chen2021EvaluatingLL}.
If models can be made honest and only assert what they believe, then they can produce outputs that are more representative and give human monitors a more accurate impression of their beliefs.\looseness=-1

\paragraph{Directions.} Researchers could create evaluation schemes that catch models being inconsistent \cite{Elazar2021MeasuringAI}, as inconsistency implies that they did not assert only what they believe. Others could also build tools to detect when models are hallucinating information \cite{lee2018hallucinations}.
To prevent models from outputting worse answers when they know better answers, researchers can concretize what it means for models to assert their true beliefs or to give the right impression.
Finally, to train more truthful models, researchers could create environments \cite{Peskov2020ItTT} or losses that incentivize models not to state falsehoods, repeat misconceptions \cite{owain2021}, or spread misinformation.\looseness=-1

\subsection{Hidden Model Functionality}

\subsubsection{Backdoors}
\paragraph{Motivation.} Machine learning systems risk carrying hidden ``backdoor'' or ``trojan'' controllable vulnerabilities. Backdoored models behave correctly and benignly in almost all scenarios, but in particular circumstances chosen by an adversary, they have been taught to behave incorrectly \cite{gu2017badnets}. %
Consider a backdoored facial recognition system that gates building access. The backdoor could be triggered by a specific unique item chosen by an adversary, such as an item of jewelry. If the adversary wears that specific item of jewelry, the backdoored facial recognition will allow the adversary into the building  \cite{sharif2016accessorize}.
A particularly important class of vulnerabilities are backdoors for sequential decision making systems, where a particular trigger leads an agent or language generation model to pursue a coherent and destructive sequence of actions \cite{Wang2020StopandGoEB, Zhang2020TrojaningLM}.

Whereas adversarial examples are created at test time, backdoors are inserted by adversaries at training time.
One way to create a backdoor is to directly inject the backdoor into a model's weights \cite{Schuster2020YouAM,Hong2021HandcraftedBI}, but they can
also be injected by adding poisoned data into the training or pretraining data \cite{Shafahi2018PoisonFT}. Injecting backdoors through poisoning is becoming easier as ML systems are increasingly trained on uncurated data scraped from online---data that adversaries can poison. If an adversary uploads a few carefully crafted poisoned images \cite{Carlini2021PoisoningAB},
code snippets \cite{Schuster2020YouAM},
or sentences \cite{Wallace2021ConcealedDP} to platforms such as Flickr, GitHub or Twitter, they can inject a backdoor into future models trained on that data \cite{Bagdasaryan2020BlindBI}.
Moreover, since downstream models are increasingly obtained by a single upstream model
\cite{Bommasani2021OnTO}, a single compromised model could 
proliferate backdoors.\looseness=-1 %

\paragraph{Directions.} To avoid deploying models that may take unexpected turns and have vulnerabilities that can be controlled by an adversary, researchers could improve backdoor detectors to combat an ever-expanding set of backdoor attacks \cite{Karra2020TheTS}. Creating algorithms and techniques for detecting backdoors is promising, but to stress test them we need to simulate an adaptive competition where researchers take the role of both attackers and auditors. This type of competition could also serve as a valuable way of grounding general hidden model functionality detection research. Researchers could try to cleanse models with backdoors, reconstruct a clean dataset given a model \cite{Yin2020DreamingTD,Wang2021IMAGINEIS}, and build techniques to detect poisoned training data. Research should also develop methods for addressing backdoors that are manually injected, not just those injected through data poisoning.\looseness=-1

\subsubsection{Emergent Hazardous Capabilities}\label{sec:emerge}
\paragraph{Motivation.} We are better able to make models safe when we know what capabilities they possess.
For early ML models, knowing their limits was often trivial, as models trained on MNIST can do little more than classify handwritten images.
However, recent large-scale models often have capabilities that their designers do not initially realize, %
with %
novel and qualitatively distinct capabilities emerging as scale increases.
For example, as GPT-3 models became larger \cite{Brown2020LanguageMA}, they gained the ability to perform arithmetic, even though GPT-3 received no explicit arithmetic supervision. 
Others have observed instances where a model's training loss remains steady, but then its test performance spontaneously ascends from random chance to perfect generalization \cite{grokking}. 
Sometimes capabilities are only discovered after initial release. %
After a multimodal image and text model \cite{Radford2021LearningTV} was released, users eventually found that its synthesized images could be markedly improved by appending ``generated by Unreal Engine'' to the query \cite{unreal}. 
Future ML models may, when prompted carefully, make the synthesis of harmful or illegal content seamless (such as videos of child exploitation, suggestions for evading the law, or instructions for building bombs). These examples demonstrate that it will be difficult to safely deploy models if we do not know their capabilities.

Some emergent capabilities may resist monitoring. In the future, it is conceivable that agent-like models may be inadvertently incentivized to adopt covert behavior. This is not unprecedented, as even simple digital organisms can evolve covert behavior. For instance, Ofria's \cite{Lehman2018TheSC} digital organisms evolved to detect when they were being monitored and would ``play dead'' to bypass the monitor, only to behave differently once monitoring completed.
In the automotive industry,
Volkswagen created products designed to bypass emissions monitors, underscoring that evading monitoring is sometimes incentivized in the real world. %
Advanced ML agents may be inadvertently incentivized to be deceptive not out of malice but simply because doing so may help maximize their human approval objective. If advanced models are also capable planners, they could be skilled at obscuring their deception from monitors.%

\paragraph{Directions.} To protect against emergent capabilities, researchers could create techniques and tools to inspect models and better foresee unexpected jumps in capabilities.
We also suggest that large research groups begin scanning models for numerous potential and as yet unobserved capabilities. We specifically suggest focusing on capabilities that could create or directly mitigate hazards. One approach is to create a continually evolving testbed to screen for potentially hazardous capabilities, such as the ability to execute malicious 
user-supplied code,
generate illegal or unethical forms of content,
or to write convincing but wrong text on arbitrary topics. Another more whitebox approach would be to predict a model's capabilities given only its weights, which might reveal latent capabilities that are not obviously expressible from standard prompts. 

Detection methods will require validation to ensure they are sufficiently sensitive. Researchers could implant hidden functionality to ensure that detection methods can detect known flaws; this can also help guide the development of 
better methods. Other directions include quantifying and extrapolating future model capabilities \cite{Henighan2020ScalingLF,Hestness2017DeepLS} and searching for novel failure modes that may be symptoms of unintended functionality.

Once a hazardous capability such as deception or illegal content synthesis is identified, the capability must be prevented or removed. Researchers could create training techniques so that undesirable capabilities are not acquired during training or during test-time adaptation. For ML systems that have already acquired an undesirable capability, researchers could create ways to teach ML systems how to forget that capability.  %
However, it may not be straightforward to determine whether the capability is truly absent and not merely obfuscated or just removed partially.

\section{Alignment}\label{sec:alignment}

\begin{figure}[H]
    \vspace{-20pt}
    \centering
    \includegraphics[width=\textwidth]{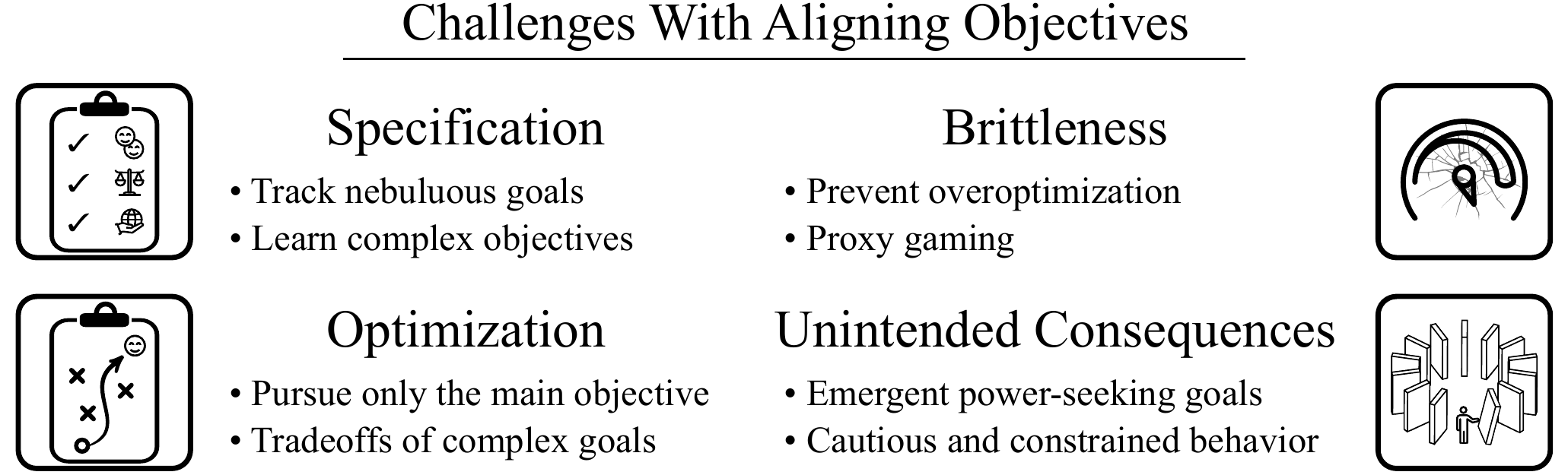}
    \vspace{-15pt}
    \caption{Alignment research aims to create and safely optimize ML system objectives.}
    \label{fig:alignment}
    \vspace{-10pt}
\end{figure}

While most technologies do not have goals and are simply tools, future machine learning systems may be more agent-like. How can we build ML agents that prefer good states of the world and avoid bad ones?
Objective functions drive system behavior, but aligning objective functions with human values requires overcoming societal as well as technical challenges. We briefly discuss societal challenges with alignment and then describe technical alignment challenges in detail.

Ensuring powerful future ML systems have aligned goals may be challenging because their goals may be given by some companies that do not solely pursue the public interest.
Unfortunately, sometimes corporate incentives can be distorted in the pursuit of maximizing shareholder value \cite{jensen1976theory}. %
Many companies help satisfy human desires and improve human welfare, but some companies have been incentivized to decimate rain forests \cite{Geist2001WhatDT}, lie to customers that cigarettes are healthy \cite{Botvin1993SmokingBO}, invade user privacy \cite{Zuboff2019TheAO}, and cut corners on safety \cite{sutton2010chromium}. %
Even if economic entities were more aligned, such as if corporations absorbed their current negative externalities, the larger economic system would still not be fully aligned with all human values.
This is because the overall activity of the economy can be viewed as approximating material wealth maximization \cite{posner}. However, once wealth increases enough, it ceases to be correlated with emotional wellbeing and happiness \cite{KahnemanDeaton}. Furthermore, wealth maximization with advanced ML may sharply exacerbate inequality \cite{greenwood1997third}, which is a robust predictor of aggression and conflict \cite{Fajnzylber2002InequalityAV}. Under extreme automation in the future, wealth metrics such as real GDP per capita may drift further from tracking our values \cite{Brynjolfsson2009WhatTG}. Given these considerations, the default economic objective shaping the development of ML is not fully aligned with human values.\looseness=-1

Even if societal issues are resolved and ideal goals are selected, technical problems remain. We focus on four important technical alignment problems: objective proxies are difficult to specify, objective proxies are difficult to optimize, objective proxies can be brittle, and objective proxies can spawn unintended consequences.

\subsection{Objectives Can Be Difficult to Specify}

\paragraph{Motivation for Value Learning.} Encoding human goals and intent is challenging. Lawmakers know this well, as laws specified by stacks of pages still often require that people interpret the spirit of the law.
Many human values, such as happiness \cite{LazariRadek2014ThePO}, good judgment \cite{Stanovich2016TheRQ}, %
meaningful experiences \cite{fbupdate}, human autonomy, and so on, are hard to define and measure.
Systems will optimize what is measurable \cite{Ridgway1956DysfunctionalCO}, as ``what gets measured gets managed.'' Measurements such as clicks and watch time may be easily measurable, but they often leave out and work against important human values such as wellbeing \cite{kross2013facebook,fbupdate,Stray2020AligningAO,Stray2021WhatAY}. Researchers will need to confront the challenge of measuring abstract, complicated, yet fundamental human values.

\paragraph{Directions.} %
Value learning seeks to develop better approximations of our values, so that corporations and policy makers can give systems better goals to pursue.
Some important values %
include wellbeing, fairness, and people getting what they deserve.
To model wellbeing, future work could use ML to model what people find pleasant, how stimuli affect internal emotional valence, and other aspects of subjective experience. Other work could try to learn how to align specific technologies, such as recommender systems, with wellbeing goals rather than engagement. Future models deployed in legal contexts must understand justice, so models should be taught the law \cite{Hendrycks2021MeasuringMM}. Researchers could create models that learn wellbeing functions that do not mimic cognitive biases \cite{Hendrycks2021AligningAW}. Others could make models that are able to detect when scenarios are clear-cut or highly morally contentious \cite{Hendrycks2021AligningAW}.  
Other directions include learning difficult-to-specify goals in interactive environments \cite{HadfieldMenell2016CooperativeIR}, learning the idiosyncratic values of different stakeholders \cite{Liao2019BuildingJC}, and learning about cosmopolitan goals such as endowing humans with the capabilities necessary for high welfare \cite{Nussbaum2003CAPABILITIESAF}.\looseness=-1

\subsection{Objectives Can Be Difficult to Optimize}

\paragraph{Motivation for Translating Values Into Action.} Putting knowledge from value learning into practice may be difficult because optimization is difficult. For example, many sparse objectives are easy to specify but difficult to optimize. Worse, some human values are particularly difficult to optimize. Take, for instance, the optimization of wellbeing. Short-term and long-term wellbeing are often anticorrelated, as the hedonistic paradox shows \cite{sidgwick_1907}. Hence many local search methods may be especially prone to bad local optima, and they may facilitate the impulsive pursuit of pleasure.
Consequently, optimization needs to be on long timescales, but this reduces our ability to test our systems iteratively and rapidly, and ultimately to make them work well. %
Further, human wellbeing is difficult to compare and trade off with other complex values, is difficult to forecast even by humans themselves \cite{Wilson2005AffectiveF}, and wellbeing often quickly adapts and thereby nullifies interventions aimed at improving it \cite{Brickman1971HedonicRA}. Optimizing complex abstract human values is therefore not straightforward.\looseness=-1

To build systems that optimize human values well, models will need to mediate their knowledge from value learning into appropriate action.
Translating background knowledge into choosing the best action is typically not straightforward: while computer vision models are advanced, successfully applying vision models for robotics remains elusive. Also, while sociopaths are intelligent and have moral awareness, this knowledge does not necessarily result in moral inclinations or moral actions.

As systems make objectives easier to optimize and break them down into new goals, subsystems are created that optimize these new intrasystem goals.  
But a common failure mode is that ``intrasystem goals come first'' \cite{Gall1977SystemanticsHS}. These goals can steer actions instead of the primary objective \cite{Hubinger2019RisksFL}. Thus a system's explicitly written objective is not necessarily the objective that the system operationally pursues, and this can result in misalignment.\looseness=-1

\paragraph{Directions.} To make models optimize desired objectives and not pursue undesirable secondary objectives, researchers could try to construct systems that guide models not just to follow rewards but also behave morally \cite{jiminy2021}; such systems could also be effective at guiding agents not to cause wanton harm within interactive environments and to abide by rules. To get a sense of an agent's values and see how it make tradeoffs between values, researchers could also create diverse environments that capture realistic morally salient scenarios and characterize the choices that agents make when faced with ethical quandaries. Research on steerable and controllable text generation \cite{Krause2020GeDiGD,Kenton2021AlignmentOL} could help chatbots exhibit virtues such as friendliness and honesty.\looseness=-1

\subsection{Objective Proxies Can Be Brittle}

Proxies that approximate our objectives are brittle, but work on Proxy Gaming and Value Clarification can help.\looseness=-1 %

\paragraph{Motivation for Proxy Gaming.} Objective proxies can be gamed by optimizers and adversaries. For example, to combat a cobra infestation, a governor of Delhi offered bounties for dead cobras. However, as the story goes, this proxy was brittle and instead incentivized citizens to breed cobras, kill them, and collect a bounty. %
In other contexts, some students overoptimize their GPA proxies by taking easier courses, and some academics overoptimize bibliometric proxies at the expense of research impact. 
Agents in reinforcement learning often find holes in proxies. In a boat racing game, an RL agent gained a high score not by finishing the race but by 
going in the wrong direction, catching on fire, and colliding into other boats \cite{boatrace}. Since proxies ``will tend to collapse once pressure is placed upon'' them by optimizers \cite{Goodhart1984ProblemsOM,Manheim2018CategorizingVO,Strathern1997ImprovingRA}, proxies can often be gamed.\looseness=-1

\begin{wrapfigure}{r}[0.01\textwidth]{.4\textwidth}%
	\vspace{-8pt}
	``When a measure becomes a target, it ceases to be a good measure.''\hfill\emph{Goodhart's Law}%
	\vspace{-5pt}
\end{wrapfigure}

\paragraph{Directions.} Advancements in robustness and monitoring are key to mitigating proxy gaming.

ML systems encoding proxies must become more robust to optimizers, which is to say they must become more adversarially robust (\Cref{sec:advex}). %
Specifically, suppose a neural network is used to define a learned utility function; if some other agent (say another neural network) is tasked with maximizing this utility proxy, it would be incentivized to find and exploit any errors in the learned
\vspace{-5pt}utility proxy, similar to adversarial examples \cite{Trabucco2021ConservativeOM,Gleave2020AdversarialPA}. Therefore we should seek to ensure adversarial robustness of learned reward functions, and regularly test them for exploitable loopholes.

Separately, advancements in monitoring can help with proxy gaming. %
For concreteness, we discuss how monitoring can specifically help with ``human approval'' proxies, but many of these directions can help with proxy gaming in general. A notable failure mode of human approval proxies is their susceptibility to deception. Anomaly detectors (\Cref{sec:anom}) could help spot when ML models are being deceptive or stating falsehoods, could help monitor agent behavior for unexpected activity, and could help determine when to stop the agent or intervene. 
Research on making models honest and teaching them to give the right impression (\Cref{sec:honest}) can help mitigate deception from models trying to game approval proxies.
To make models more truthful and catch deception, future systems could attempt to verify statements that are difficult for humans to check in reasonable timespans, and they could inspect convincing but not true assertions \cite{Peskov2020ItTT}. Researchers could determine the veracity of model assertions, possibly through an adversarial truth-finding process \cite{Irving2018AISV}.

\paragraph{Motivation for Value Clarification.} While maximization can expose faults in proxies, so too can future events. The future will sharpen and force us to confront unsolved ethical questions %
about our values and objectives \cite{Williams2015ThePO}. In recent decades, peoples' values have evolved by confronting philosophical questions, including whether to infect volunteers for science, how to equitably distribute vaccines,
the rights of people with different orientations, %
and so on. How are we to act if many humans spend  most of their time chatting with compelling bots and not much time with humans, %
or how should we fairly address automation's economic ramifications? Determining the right action is not strictly scientific in scope \cite{hume}, and we will need philosophical analysis to help us correct structural faults in our proxies.\looseness=-1

\paragraph{Directions.} %
We should build systems to help rectify our objectives and proxies, so that we are less likely to optimize the wrong objective when a change in goals is necessary. This requires interdisciplinary research towards a system that can reason about values and philosophize at an expert level.
Research could start with trying to build a system to score highly in the philosophy olympiad, in the same way others are aiming to build expert-level mathematician systems using mathematics olympiad problems \cite{Maric2020FormalizingIP}. Other work could build systems to help extrapolate the end products of ``reflective equilibrium'' \cite{rawls}, or what objectives we would endorse by simulating a process of deliberation about competing values.
Researchers could also try to estimate the quality of a philosophical work by using a stream of historical philosophy papers and having models predict the impact of each paper on the literature.
Eventually, researchers should seek to build systems that can formulate robust positions through an argumentative dialog. These systems could also try to find flaws in verbally specified proxies.\looseness=-1 %

\subsection{Objective Proxies Can Lead to Unintended Consequences}

\paragraph{Motivation.} While optimizing agents may work towards subverting a proxy, in other situations both the proxy setter and an optimizing agent can fall into states that neither intended. %
For example, in their pursuit to modernize the world with novel technologies, previous well-intentioned scientists and engineers inadvertently increased pollution and hastened climate change, an outcome desired neither by the scientists themselves nor by the societal forces that supported them. %
In ML, some platforms maximized clickthrough rates to approximate maximizing enjoyment, but such platforms unintentionally addicted many users and decreased their wellbeing. These cases demonstrate that unintended consequences present a challenging but important problem.\looseness=-1 %

\paragraph{Directions.} Future research could focus on designing minimally invasive agents that prefer easily reversible to irreversible actions \cite{Grinsztajn2021ThereIN}, as irreversibility reduces humans' optionality and often unintentionally destroys potential future value. Likewise, researchers could create agents that properly account for their lack of knowledge of the true objective \cite{HadfieldMenell2017TheOG} and avoid disrupting parts of the environment whose value is unclear \cite{Turner2020AvoidingSE,Krakovna2020AvoidingSE,Shah2019PreferencesII}. We also need more complex environments that can manifest diverse unintended side effects \cite{Wainwright2020SafeLife1E} such as feedback loops, which are a source of hazards to users of recommender systems \cite{Krueger2020HiddenIF}. A separate way to mitigate unintended consequences is to teach ML systems to abide by constraints \cite{Achiam2019BenchmarkingSE,Saunders2018TrialWE}, be less brazen, and act cautiously. Since we may be uncertain about which values are best, research could focus on having agents safely optimize and balance many values, so that one value does not unintentionally dominate or subvert the rest \cite{Newberry2021ThePA,Ecoffet2021ReinforcementLU}. Sometimes unintended instrumental goals emerge in systems, such as self-preservation \cite{HadfieldMenell2017TheOG} or power-seeking \cite{turner2021optimal}, so researchers %
could try mitigating and detecting such unintended emergent goals; see \Cref{sec:emerge} for more directions in detecting emergent functionality.\looseness=-1

\section{Systemic Safety}
\begin{figure}[H]
    \centering
    \includegraphics[width=0.78\textwidth]{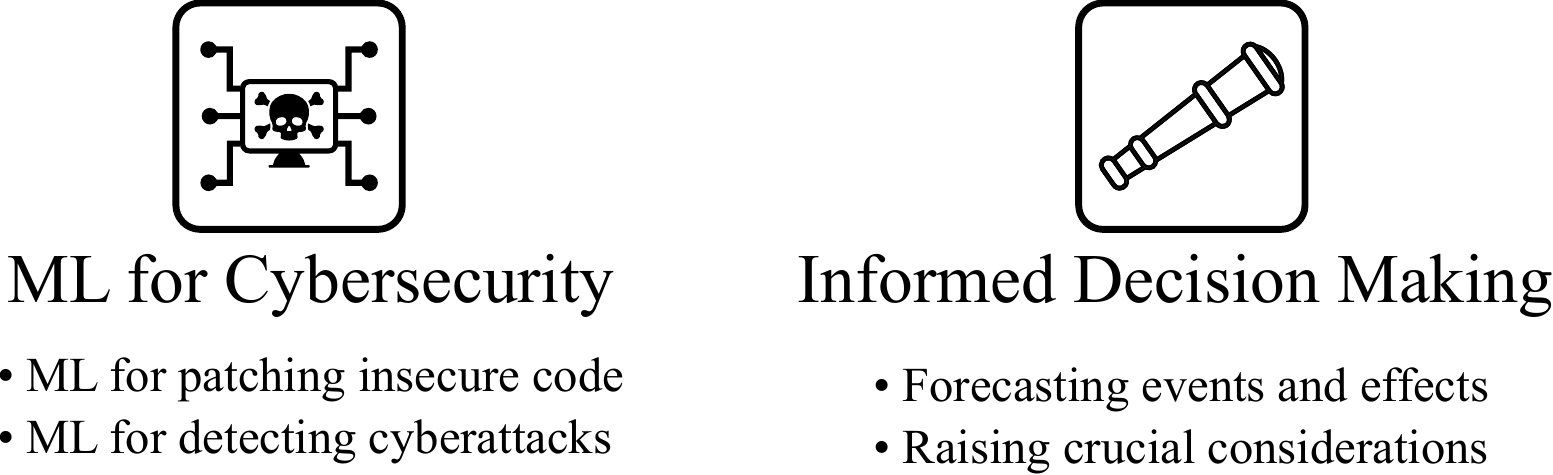}
    \caption{Systemic safety research aims to address broader contextual risks to how ML systems are handled. Both cybersecurity and decision making may decisively affect whether ML systems will fail or be misdirected. %
    }
    \label{fig:externalsafety}
\end{figure}

Machine learning systems do not exist in a vacuum, and the safety of the larger context can influence how ML systems are handled and affect the overall safety of ML systems. ML systems are more likely to fail or be misdirected if the larger context in which they operate is insecure or turbulent. %

Systemic safety research applies ML to mitigate potential contextual hazards that may decisively cause ML systems to fail or be misdirected. As two examples, we support research on cybersecurity and on informed decision making. The first problem is motivated by the observation that ML systems are integrated with vulnerable software, and in the future ML may change the landscape of cyberattacks.
In the second problem, we turn to a speculative approach for improving governance decisions and command and control operations using ML, as institutions may direct the most powerful future ML systems.

Beyond technical work, policy and governance work will be integral to safe deployment \cite{dafoe2018ai,Bender2021OnTD,Birhane2021TheVE,zwetsloot2018beyond,Brundage2020TowardTA}. While techno-solutionism has limitations, technical ML researchers should consider using their skillset to address deployment environment hazards, and we focus on empirical ML research avenues, as we expect most readers are technical ML researchers.

Finally, since there are multiple hazards that can hinder systemic safety, this section is nonexhaustive. For instance, if ML industry auditing tools could help regulators more effectively regulate ML systems, research developing such tools could become part of systemic safety. Likewise, using ML to help facilitate cooperation~\cite{Dafoe2020OpenPI} may emerge as a research area.

\subsection{ML for Cybersecurity}
\paragraph{Motivation.} Cybersecurity risks can make ML systems unsafe, as ML systems operate in tandem with traditional software and are often instantiated as a cyber-physical system. As such, malicious actors could exploit insecurities in traditional software to control autonomous ML systems. Some ML systems may also be private or unsuitable for proliferation, and they will therefore need to operate on computers that are secure.

Separately, ML may amplify future automated cyberattacks and enable malicious actors to increase the accessibility, potency, success rate, scale, speed, and stealth of their attacks. For example, hacking currently requires specialized skills, but if state-of-the-art ML models could be fine-tuned for hacking, then the barrier to entry for hacking may decrease sharply.
Since cyberattacks can destroy valuable information and even destroy critical physical infrastructure \cite{Cary2020DestructiveCO} such as power grids \cite{Ottis2008AnalysisOT} and building hardware \cite{Langner2011StuxnetDA}, these potential attacks are a looming threat to international security.

While cybersecurity aims to increase attacker costs, the cost-benefit analysis may become lopsided if attackers eventually gain a larger menu of options that require negligible effort. In this new regime, attackers may gain the upper hand, like how attackers of ML systems currently have a large advantage over defenders.
Since there may be less of a duality between offensive and defensive security in the future, we suggest that research focus on techniques that are clearly defensive. %
The severity of this risk is speculative, but neural networks are now rapidly gaining the ability to write code and interact with the outside environment, and at the same time there is very little research on deep learning for cybersecurity. %

\paragraph{Directions.} To mitigate the potential harms of automated cyberattacks to ML and other systems, researchers should apply ML to develop better defensive techniques.
For instance, ML could be used to detect intruders \cite{lane1997application,sommer2010outside} or impersonators \cite{Ho2019DetectingAC}. ML could also help analyze code and detect software vulnerabilities.%
Massive unsupervised ML methods could also model binaries and learn to detect malicious obfuscated payloads \cite{sgn,Shin2015RecognizingFI,ghidra,harang2020sorel20m}. %
Researchers could also create ML systems that model software behavior and detect whether programs are sending packets when they should not. %
ML models could help predict future phases of cyberattacks, and such automated warnings could be judged by their lead time, precision, recall, and the quality of their contextualized explanation. Advancements in code translation \cite{Lachaux2020UnsupervisedTO,Austin2021ProgramSW} and code generation \cite{Chen2021EvaluatingLL,Pearce2021AnEC} suggest that future models could apply security patches and make code more secure, so that future systems not only flag security vulnerabilities but also fix them. %

\subsection{Improved Epistemics and Decision Making}
\paragraph{Motivation.} Even if we create reliable ML systems, these systems will not exhibit or ensure safety if the institutions that steer ML systems make poor decisions. Although nuclear weapons are a reliable and dependable technology, they became especially unsafe during the Cold War. During that time, misunderstanding and political turbulence exposed humanity to several close calls and brought us to the brink of catastrophe, demonstrating that systemic safety issues can make technologies unsafe. 
The most pivotal decisions are made during times of crisis, and future crises may be similarly risky as ML continues to be weaponized \cite{lethalaw,OpenLetter}. 
This is why we suggest creating tools to help decision-makers handle ML systems in highly uncertain, quickly evolving, turbulent situations.\looseness=-1

\paragraph{Directions.} To improve the decision-making and epistemics of political leaders and command and control centers, 
we suggest two efforts: using ML to improve forecasting and bringing to light crucial considerations.

Many governance and command and control decisions are based on forecasts \cite{Tetlock2015SuperforecastingTA} from humans, and some forecasts are starting to incorporate ML \cite{gide3}. Forecasters assign probabilities to possible events that could happen within the next few months or years (e.g., geopolitical, epidemiological, and industrial events), and are scored by their correctness and calibration. To be successful, forecasters must dynamically aggregate information from disparate unstructured sources \cite{Jin2021ForecastQAAQ}. 
This is challenging even for humans, but ML systems could potentially aggregate more information, be faster, be nonpartisan, consider multiple perspectives, and thus ultimately make more accurate predictions \cite{integrativecomplexity}. The robustness of such systems could be assessed based on their ability to predict pivotal historical events, if the model only has access to data before those events. An accurate forecasting tool would need to be applied with caution to prevent over-reliance \cite{Hedlund82}, and it would need to present its data carefully so as not to encourage risk-taking behavior from the humans operating the forecasting system \cite{Taleb2013OnTD}.

Separately, researchers should develop systems that identify questions worth asking and crucial factors to consider. While forecasting can refine estimates of well-defined risks, these advisory systems could help unearth new sources of risk and identify actions to mitigate risks. Since ML systems can process troves of historical data and can learn from diverse situations during training, they could suggest possibilities that would otherwise require extensive memory and experience. Such systems could help orient decision making by providing related prior scenarios and relevant statistics such as base rates.
Eventually advisory systems could identify stakeholders, propose metrics, brainstorm options, suggest alternatives, and note trade-offs to further improve decision quality \cite{Gathani2021AugmentingDM}.
In summary, ML systems that can predict a variety of events and identify crucial considerations could help provide good judgment and correct misperceptions, and thereby reduce the chance of rash decisions and inadvertent escalation.\looseness=-1

\begin{figure*}[t]
    \centering
    \includegraphics[width=\textwidth]{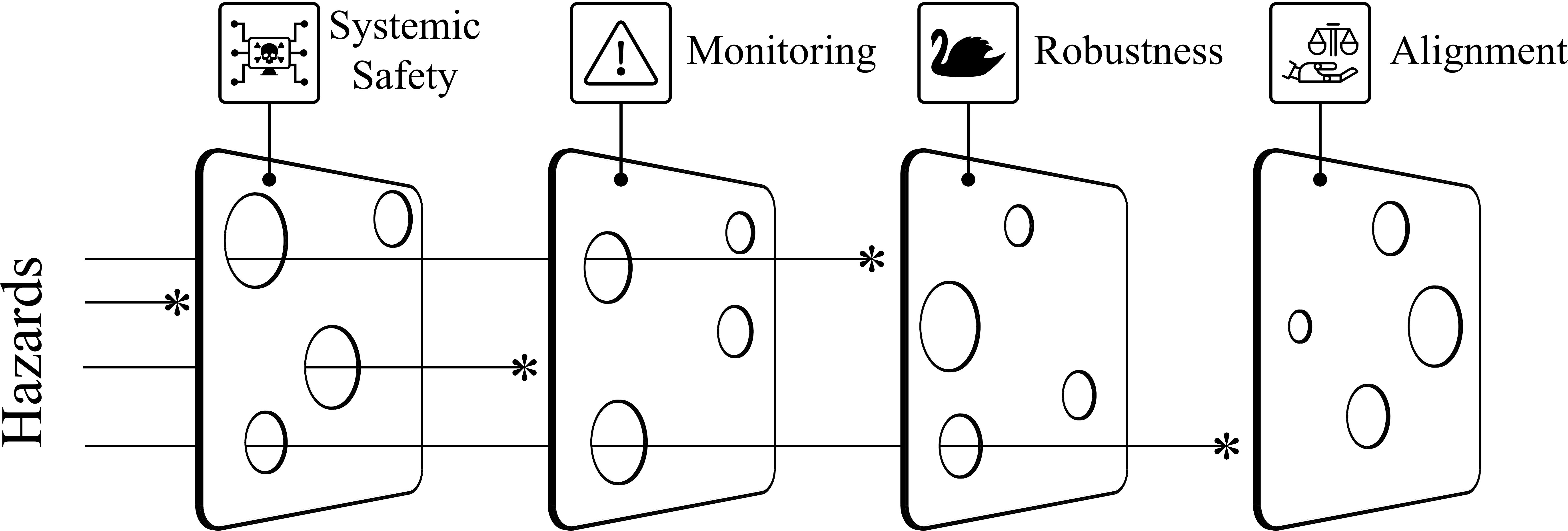}
    \caption{A Swiss cheese model of ML Safety research. Pursuing multiple safety research avenues creates multiple layers of protection which mitigates hazards and makes ML systems safer.}
    \label{fig:swiss}
\end{figure*}

\vspace{3pt}
\section{Related Research Agendas}
\vspace{3pt}
There is a large ecosystem of work on addressing societal consequences of machine learning, including AI policy \cite{dafoe2018ai}, privacy \cite{Abadi2016DeepLW,shokri2017membership}, fairness \cite{Hardt2016EqualityOO}, and ethics \cite{Gabriel2020ArtificialIV}. %
We strongly support research on these related areas. %
For purposes of scope, in this section we focus on papers that outline paths towards creating safe ML systems.\\

An early work that helps identify safety problems is Russell \emph{et al.}, 2015 \cite{Russell2015ResearchPF}, who identify many potential avenues for safety, spanning robustness, machine ethics, research on AI's economic impact, and more. Amodei and Olah \emph{et al.}, 2016 \cite{Amodei2016ConcretePI} helped further concretize several safety research directions. With the benefit of five years of hindsight, %
our paper provides a revised and expanded collection of concrete problems. %
Some of our themes extend the themes in Amodei and Olah \emph{et al.}, such as Robustness and some portions of Alignment. We focus here on problems that remain unsolved and also identify new problems, such as emergent capabilities from massive pretrained models, that stem from recent progress in ML. We also broaden the scope by identifying systemic safety risks surrounding the deployment context of ML. %
The technical agenda of Taylor \emph{et al.}, 2016 \cite{Taylor2016AlignmentFA} considers similar topics to Amodei and Olah \emph{et al.}, and Leike \emph{et al.}, 2018 \cite{Leike2018ScalableAA} considers safety research directions in reward modeling. Although Leike \emph{et al.}'s research agenda focuses on reinforcement learning, they highlight the importance of various other research problems including adversarial training and uncertainty estimation.
Recently, Critch and Krueger, 2020 \cite{Critch2020AIRC} provide an extensive commentary on safety research directions and discuss safety when there are multiple stakeholders.\looseness=-1

\section{Conclusion}
This work presented a non-exhaustive list of four unsolved research problems, all of which are interconnected and interdependent.
Anomaly detection, for example, helps with detecting proxy gaming, detecting suspicious cyberactivity, and executing fail-safes in the face of unexpected events. Achieving safety requires research on all four problems, not just one. To see this, recall that a machine learning system that is not aligned with human values may be unsafe in and of itself, as it may create unintended consequences or game human approval proxies. Even if it is possible to create aligned objectives for ML systems, Black Swan events could cause ML systems to misgeneralize and pursue incorrect goals, malicious actors may launch adversarial attacks or compromise the software on which the ML system is running, and humans may need to monitor for emergent functionality and the malicious use of ML systems. As depicted in \Cref{fig:swiss}'s highly simplified model, work on all four problems helps create comprehensive and layered protective measures against a wide range of safety threats.\looseness=-1 %

As machine learning research evolves, the community's aims and expectations should evolve too.
For many years, the machine learning community focused on making machine learning systems work in the first place. %
However, machine learning systems have had notable success in domains from images, to natural language, to programming---therefore our focus should expand beyond just accuracy, speed, and scalability. Safety must now become a top priority. %

Safety is not auxiliary in most current widely deployed technology. Communities do not ask for ``safe bridges,'' but rather just ``bridges.''  Their safety is insisted upon---even assumed---and incorporating safety features is imbued in the design process. The ML community should similarly create a culture of safety and elevate its standards so that ML systems can be deployed in safety-critical situations.

\newpage
\subsection*{Acknowledgements}
We would like to thank Sidney Hough, Owain Evans, Collin Burns, Alex Tamkin, Mantas Mazeika, Kevin Liu, Jonathan Uesato, Steven Basart, Henry Zhu, D. Sculley, Mark Xu, Beth Barnes, Andreas Terzis, Florian Tram\`er, Stella Biderman, Leo Gao, Jacob Hilton, and Thomas Dietterich for their feedback. DH is supported by the NSF GRFP Fellowship and an Open Philanthropy Project AI Fellowship.

\printbibliography
\newpage
\begin{appendices}
\crefalias{section}{appsec}

\section{Analyzing Risks, Hazards, and Impact}

\subsection{Risk Management Framework}

\begin{table}[H]
\setlength{\tabcolsep}{6pt}
\setlength\extrarowheight{2pt}
\centering
\begin{tabularx}{1.0\textwidth}{*{1}{>{\hsize=0.85\hsize}X} *{1}{>{\hsize=4.8cm}X }
|  *{1}{>{\hsize=0.9\hsize}Y} *{1}{>{\hsize=0.95\hsize}Y} *{1}{>{\hsize=0.95\hsize}Y} *{1}{>{\hsize=0.55\hsize}Y}
}
\thead[l]{Area} & \thead[l]{Problem} & \thead{ML\\System\\Risks} & \thead{Operational\\Risks} & \thead{Institutional\\ and Societal\\Risks} & \thead{Future\\Risks} \\ \hline
 \parbox[t]{50mm}{\multirow{2}{*}{Robustness}}
 & Black Swans and Tail Risks  & \checkmark & \checkmark &  &  \checkmark\\
 & Adversarial Robustness & \checkmark &  &  & \checkmark \\
 \Xhline{0.5\arrayrulewidth} 
 \parbox[t]{50mm}{\multirow{3}{*}{Monitoring}}
 & Anomaly Detection  & \checkmark & \checkmark &  & \checkmark \\
 & Representative Outputs  & \checkmark & \checkmark &  & \checkmark \\
 & Hidden Model Functionality  & \checkmark & \checkmark &  & \checkmark  \\
 \Xhline{0.5\arrayrulewidth} 
 \parbox[t]{50mm}{\multirow{5}{*}{Alignment}}
 & Value Learning    &  & \checkmark & \checkmark & \checkmark \\
 & Translating Values to Action    & \checkmark &  &  & \checkmark \\
 & Proxy Gaming   & \checkmark &  &  & \checkmark \\
 & Value Clarification    &  &  & \checkmark & \checkmark \\
 & Unintended Consequences    & \checkmark & \checkmark & \checkmark & \checkmark \\
 \Xhline{0.5\arrayrulewidth} 
Systemic & ML for Cybersecurity    & \checkmark & \checkmark & \checkmark & \checkmark  \\
Safety & Informed Decision Making  &  & \checkmark & \checkmark & \checkmark \\
 \bottomrule
\end{tabularx}
\caption{Problems and the risks they directly mitigate. Each checkmark indicates whether a problem directly reduces a risk. Notice that problems affect both near- and long-term risks. %
}
\label{tab:problemsandrisks}
\end{table}

To analyze how ML Safety progress can reduce abstract risks and hazards,\footnote{%
One can think of hazards as factors that have the potential to cause harm. One can think of risk as the hazard's prevalence multiplied by the amount of exposure to the hazard multiplied by the hazard's deleterious effect. For example, a wet floor is a hazard to humans. However, risks from wet floors are lower if floors dry more quickly with a fan (systemic safety). Risks are lower if humans heed wet floor signs and have less exposure to them (monitoring). Risks are also lower for young adults than the elderly, since the elderly are more physically vulnerable (robustness).
In other terms, robustness makes systems less vulnerable to hazards, monitoring reduces exposure to hazards, alignment makes systems inherently less hazardous, and systemic safety reduces systemic hazards.} we identify four dimensions of risk in this section and five hazards in the next section.

The following four risk dimensions are adopted from the Department of Defense's broad risk management framework \cite{DoD}, with its personnel management risks replaced with ML system risks.

\begin{enumerate}
\item \textbf{ML System Risks} -- risks to the ability of a near-term individual ML system to operate reliably.
\item \textbf{Operational Risks} -- risks to the ability of an organization to safely operate an ML system in near-term deployment scenarios.
\item \textbf{Institutional and Societal Risks} -- risks to the ability of global society or institutions that decisively affect ML systems to operate in near-term scenarios in an efficient, informed, and prudent way.
\item \textbf{Future (ML System, Operational, and Institutional) Risks} -- risks to the ability of future ML systems, organizations operating ML systems, and institutions to address mid- to long-term challenges.
\end{enumerate}

In \Cref{tab:problemsandrisks}, we indicate whether one of these risks is reduced by progress on a given ML Safety problem.
Note that these all problems reduce risks to all three of future ML systems, organizations, and institutions. In the future, organizations and institutions will likely become more dependent on ML systems, so improvements to Black Swans robustness would in the future help improve operations and institutions dependent on ML systems. Since this table is a snapshot of the present, risk profiles will inevitably change.

\subsection{Hazard Management Framework}

\begin{table}[H]
\fontsize{9}{11}\selectfont
\setlength{\tabcolsep}{5pt}
\setlength\extrarowheight{2pt}
\centering
\begin{tabularx}{1.0\textwidth}{*{1}{>{\hsize=0.7\hsize}X} *{1}{>{\hsize=4.3cm}X }
|  *{1}{>{\hsize=0.75\hsize}Y} *{1}{>{\hsize=0.75\hsize}Y}  *{1}{>{\hsize=0.87\hsize}Y} *{1}{>{\hsize=0.4\hsize}Y} *{1}{>{\hsize=0.87\hsize}Y} }
\thead[l]{Area} & \thead[l]{Problem} & \thead{Known\\Unknowns} & \thead{Unknown\\Unknowns} & \thead{Emergence} & \thead{Long\\Tails} & \thead{Adversaries\\\& Deception} \\ \hline
 \parbox[t]{50mm}{\multirow{2}{*}{Robustness}}
 & Black Swans and Tail Risks  & \textcolor{rightgreen}{\ding{52}} %
 & \textcolor{rightgreen}{\ding{52}} & \checkmark & \textcolor{rightgreen}{\ding{52}} &  \\
 & Adversarial Robustness & \checkmark &  &  &  & \textcolor{rightgreen}{\ding{52}} \\
 \Xhline{0.5\arrayrulewidth} 
 \parbox[t]{50mm}{\multirow{3}{*}{Monitoring}}
 & Anomaly Detection  & \checkmark & \textcolor{rightgreen}{\ding{52}} & \textcolor{rightgreen}{\ding{52}} & \textcolor{rightgreen}{\ding{52}} & \textcolor{rightgreen}{\ding{52}} \\
 & Representative Outputs  & \textcolor{rightgreen}{\ding{52}} & \checkmark &  &  & \textcolor{rightgreen}{\ding{52}} \\
 & Hidden Model Functionality  & \textcolor{rightgreen}{\ding{52}} & \textcolor{rightgreen}{\ding{52}} & \textcolor{rightgreen}{\ding{52}} &  & \textcolor{rightgreen}{\ding{52}} \\
 \Xhline{0.5\arrayrulewidth} 
 \parbox[t]{50mm}{\multirow{5}{*}{Alignment}}
 & Value Learning  & \textcolor{rightgreen}{\ding{52}} &  &  &  &  \\
 & Translating Values to Action  & \textcolor{rightgreen}{\ding{52}} &  &  & \checkmark &  \\
 & Proxy Gaming  & \checkmark &  &  &  & \textcolor{rightgreen}{\ding{52}} \\
 & Value Clarification  & \textcolor{rightgreen}{\ding{52}} & \textcolor{rightgreen}{\ding{52}} & \checkmark & \checkmark &  \\
 & Unintended Consequences  &  & \textcolor{rightgreen}{\ding{52}} & \textcolor{rightgreen}{\ding{52}} &  \\
 \Xhline{0.5\arrayrulewidth} 
 Systemic & ML for Cybersecurity  & \checkmark & \checkmark & &  & \textcolor{rightgreen}{\ding{52}} \\
 Safety & Informed Decision Making  & \textcolor{rightgreen}{\ding{52}} & \textcolor{rightgreen}{\ding{52}} & \checkmark & \checkmark &  \\
 \bottomrule
\end{tabularx}
\caption{Problems and the hazards they help handle. Checkmarks indicate whether a problem directly reduces vulnerability or exposure to a given hazard, and bold green checkmarks indicate an especially notable reduction.}
\label{tab:problemsandhazards}
\end{table}

We now turn from what is affected by risks to five abstract hazards that create risks.\begin{wrapfigure}{R}{0.23\textwidth}
    \vspace{-22pt}
	\begin{center}
	\includegraphics[width=0.23\textwidth]{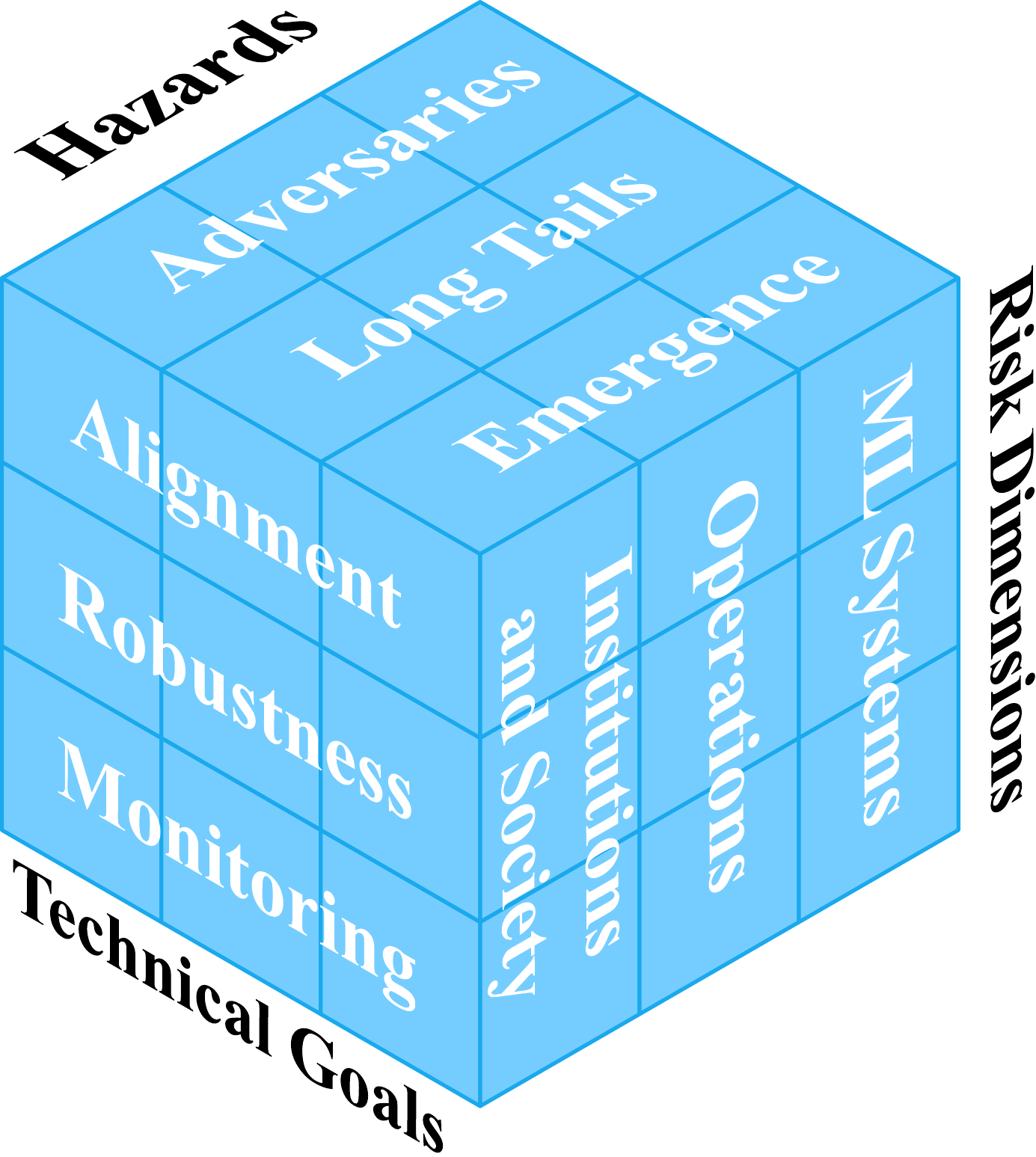}
	\end{center}
    \caption{A simplified model of interconnected factors for ML Safety.}
	\label{fig:cube}
	\vspace{-20pt}
\end{wrapfigure}

\begin{enumerate}
\item \textbf{Known Unknowns} -- Identified hazards for which we have imperfect or incomplete knowledge. These are identified hazards known to have unknown aspects. %
\item \textbf{Unknown Unknowns} -- Hazards which are unknown and unidentified, and they have properties that are unknown. %
\item \textbf{Emergence} -- A hazard that forms and comes into being as the system increases in size or its parts are combined. Such hazards do not exist in smaller versions of the system nor in its constituent parts.
\item \textbf{Long Tails} -- Hazards that can be understood as unusual or extreme events from a long tail distribution.
\item \textbf{Adversaries \& Deception} -- Hazards from a person, system, or force that aims to attack, subvert, or deceive.
\end{enumerate}

These hazards do not enumerate all possible hazards.
For example, the problems in Systemic Safety help with turbulence hazards. Furthermore, feedback loops, which can create long tails, could become a more prominent hazard in the future when ML systems are integrated into more aspects of our lives.

The five hazards have some overlap. For instance, when something novel emerges, it is an unknown unknown. When it is detected, it can become a known unknown. Separately, long tail events are often but not necessarily unknown unknowns: the 1987 stock market crash was a long tail event, but it was a known unknown to a prescient few and an unknown unknown to most everybody else. Emergent hazards sometimes co-occur with long tailed events, and an adversarial attack can cause long tail events. 

In \Cref{tab:problemsandhazards}, we indicate whether an ML Safety problem reduces vulnerability or exposure to a given hazard. As with \Cref{tab:problemsandrisks}, the table is a snapshot of the present. For example, future adversaries could create novel unusual events or strike during tail events, so Black Swan robustness could improve adversarial robustness.

With risks, hazards, and goals now all explicated, we depict their interconnectedness in \Cref{fig:cube}.

\newpage
\subsection{Prioritization and Strategy for Maximizing Impact}

\begin{table}[ht]
\setlength\extrarowheight{2pt}
\centering
\begin{tabularx}{\textwidth}{*{1}{>{\hsize=1\hsize}X} *{1}{>{\hsize=5.2cm}X }
|  *{1}{>{\hsize=1\hsize}X} *{1}{>{\hsize=1\hsize}X} *{1}{>{\hsize=1\hsize}X}}
Area & \multicolumn{1}{l|}{Problem} &
{Importance} & {Neglectedness} & {Tractability} \\ \hline
 \parbox[t]{50mm}{\multirow{2}{*}{Robustness}}
 & Black Swans and Tail Risks  & $\bullet$ $\bullet$ & $\bullet$ $\bullet$ & $\bullet$ $\bullet$ \\
 & Adversarial Robustness & $\bullet$ $\bullet$ & $\bullet$ & $\bullet$ $\bullet$ \\
 \Xhline{0.5\arrayrulewidth} 
 \parbox[t]{50mm}{\multirow{3}{*}{Monitoring}}
 & Anomaly Detection  & $\bullet$ $\bullet$ $\bullet$ & $\bullet$ $\bullet$ & $\bullet$ $\bullet$ $\bullet$ \\
 & Representative Outputs  & $\bullet$ $\bullet$ $\bullet$  & $\bullet$ $\bullet$ & $\bullet$ $\bullet$ \\
 & Hidden Model Functionality  & $\bullet$ $\bullet$ $\bullet$ & $\bullet$ $\bullet$ & $\bullet$ $\bullet$ \\
 \Xhline{0.5\arrayrulewidth} 
 \parbox[t]{50mm}{\multirow{5}{*}{Alignment}}
 & Value Learning  & $\bullet$ $\bullet$ $\bullet$ & $\bullet$ $\bullet$  & $\bullet$ $\bullet$ \\
 & Translating Values to Action  & $\bullet$ $\bullet$ & $\bullet$ $\bullet$ & $\bullet$ $\bullet$ $\bullet$ \\
 & Proxy Gaming  & $\bullet$ $\bullet$ $\bullet$ & $\bullet$ $\bullet$ & $\bullet$ $\bullet$ \\
 & Value Clarification  & $\bullet$ $\bullet$ & $\bullet$ $\bullet$ $\bullet$ & $\bullet$ \\
 & Unintended Consequences & $\bullet$ $\bullet$ & $\bullet$ $\bullet$ $\bullet$ & $\bullet$ \\
 \Xhline{0.5\arrayrulewidth} 
Systemic & ML for Cybersecurity  & $\bullet$ $\bullet$ & $\bullet$ $\bullet$ $\bullet$ & $\bullet$ $\bullet$ $\bullet$ \\
Safety & Informed Decision Making  & $\bullet$ $\bullet$ & $\bullet$ $\bullet$ & $\bullet$ $\bullet$ \\
 \bottomrule
\end{tabularx}
\caption{Problems and three factors that influence expected marginal impact.}
\label{tab:intframework}
\end{table}

We presented several problems, but new researchers may be able to make a larger impact on some problems over others. Some problems may be important, but if they are extremely popular, the risk of scooping increases, as does the risk of researchers stepping on each others' toes. Likewise, some problems may be important and may be decisive for safety if solved, but some problems are simply infeasible. Consequently, we should consider the importance, neglectedness, and tractability of problems.

\begin{enumerate}
    \item \textbf{Importance} -- How much potential risk does substantial progress on this problem reduce?
    \begin{enumerate}[leftmargin=2\parindent,align=left,labelwidth=\parindent,labelsep=10pt]
        \item[$\bullet$\phantom{ $\bullet$ $\bullet$}] Progress on this problem reduces risks of  catastrophes.
        \item[$\bullet$ $\bullet$\phantom{ $\bullet$}] Progress on this problem directly reduces risks from potential permanent catastrophes.
        \item[$\bullet$ $\bullet$ $\bullet$] Progress on this problem directly reduces risks from  more plausible permanent catastrophes.
    \end{enumerate}
    \item \textbf{Neglectedness} --  How much research is being done on the problem?
    \begin{enumerate}[leftmargin=2\parindent,align=left,labelwidth=\parindent,labelsep=10pt]
        \item[$\bullet$\phantom{ $\bullet$ $\bullet$}] The problem is one of the top ten most researched topics at leading conferences.
        \item[$\bullet$ $\bullet$\phantom{ $\bullet$}] The problem receives some attention at leading ML conferences, or adjacent problems are hardly neglected.
        \item[$\bullet$ $\bullet$ $\bullet$] The problem has few related papers consistently published at leading ML conferences.
    \end{enumerate}
    \item \textbf{Tractability} -- How much progress can we expect on the problem?
    \begin{enumerate}[leftmargin=2\parindent,align=left,labelwidth=\parindent,labelsep=10pt]
        \item[$\bullet$\phantom{ $\bullet$ $\bullet$}] We cannot expect large research efforts to highly fruitful currently, possibly due to conceptual bottlenecks, or productive work on the problem likely requires far more advanced ML capabilities.
        \item[$\bullet$ $\bullet$\phantom{ $\bullet$}] We expect to reliably and continually make progress on the problem.
        \item[$\bullet$ $\bullet$ $\bullet$] A large research effort would be highly fruitful and there is obvious low-hanging fruit.
    \end{enumerate}
\end{enumerate}

A snapshot of each problem and its current importance, neglectedness, and tractability is in \Cref{tab:intframework}. Note this only provides a rough sketch, and it has limitations. For example, a problem that is hardly neglected overall may still have neglected aspects; while adversarial robustness is less neglected than other safety problems, robustness to unforeseen adversaries is fairly neglected. Moreover, working on popular shovel-ready problems may be more useful for newcomers compared to working on problems where conceptual bottlenecks persist. Further, this gives a rough sense of marginal impact, but entire community should not chose to act in the same way marginally, or else neglected problems will suddenly become overcrowded.

These three factors are merely prioritization factors and do not define a strategy. Rather, a potential strategy for ML Safety is as follows.
\begin{enumerate}
    \item Force Management: Cultivate and maintain a force of ready personnel to implement safety measures into advanced ML systems and operate ML systems safely.
    \item Research: Build and maintain a community to conduct safety research, including the identification of potential future hazards, clarification of safety goals, reduction of the costs to adopt safety methods, research on how to incorporate safety methods into existing ML systems, and so on.
    \item Protocols: Establish and incentivize adherence to protocols, precedents, standards, and research expectations such as red teaming, all for the safe development and deployment of ML systems.
    \item Partnerships: Build and maintain safety-focused alliances and partnerships among academe, industry, and government.
\end{enumerate}

In closing, throughout ML Safety's development we have seen numerous proposed strategies, hazards, risks, scenarios, and problems. In safety, some previously proposed problems have been discarded, and some new problems have emerged, just as in the broader ML community. Since no individual knows what lies ahead, safety analysis and strategy will need to evolve and adapt beyond this document. Regardless of which particular safety problems turn out to be the most or least essential, the success of safety's evolution and adaptation rests on having a large and capable research community.

\end{appendices}

\end{document}